\documentclass[journal]{IEEEtran}
\usepackage[pdftex]{graphicx}

\ifCLASSINFOpdf
\else
\fi
\begin{document}
%
\title{Technical Report for Valence-Arousal Estimation on Affwild2 Dataset}
%
%
%

\author{I-Hsuan~Li
\thanks{M. Shell was with the Department
of Electrical and Computer Engineering, Georgia Institute of Technology, Atlanta,
GA, 30332 USA e-mail: (see http://www.michaelshell.org/contact.html).}
\thanks{J. Doe and J. Doe are with Anonymous University.}
\thanks{Manuscript received April 19, 2005; revised August 26, 2015.}}

\maketitle

\begin{abstract}
In this work, we describe our method for tackling the valence-arousal estimation challenge from ABAW FG-2020 Competition. The competition organizers provide an in-the-wild Aff-Wild2 dataset for participants to analyze affective behavior in real-life settings. We use MIMAMO Net \cite{deng2020mimamo} model to achieve information about micro-motion and macro-motion for improving video emotion recognition and achieve Concordance Correlation Coefficient (CCC) of 0.415 and 0.511 for valence and arousal on the reselected validation set.
\end{abstract}


%

\section{Introduction}
%
%
%
%
\IEEEPARstart{P}{eople} study facial expression recognition research for a long history, but still have some challenges to be addressed, such as in-the-wild dataset improvements. No in-the-wild dataset contains complete and various sets of labels for human emotion estimation, due to the high cost and time requirement. And this causes the limitation of multi-task method progression and applications in real life. Recently, to tackle such problems, Kollias et al. \cite{kollias2020analysing}, \cite{kollias2019expression}, \cite{kollias2019face}, \cite{zafeiriou2017aff}, \cite{kollias2017recognition} held Affective Behavior Analysis in-the-wild (ABAW) FG-2020 Competition and built the large-scale Aff-Wild2 dataset, which includes annotations of valence/arousal value, action unit (AU), and facial expression for three different recognition tasks.
Valence represents how positive the person is while arousal describes how active the person is. AUs are the basic actions of individuals or groups of muscles for portraying emotions. As for facial expression, it classifies into seven categories, neutral, anger, disgust, fear, happiness, sadness, and surprise.

The challenges for ABAW FG-2020 Competition include valence-arousal estimation, facial action unit detection, and expression classification. We focus on valence-arousal estimation by using MIMAMO Net, which was proposed by Deng et al \cite{deng2020mimamo}. The model, shown in Fig. 1, gets state-of-the-art performance on the OMG \cite{barros2018omg} and Aff-Wild \cite{kollias2019deep} dataset. It uses spatial-temporal feature learning to capture information about micro-motion and macro-motion and combine them by GRU network to improve video emotion recognition.

\section{RELATED WORK}
In recent years, most of the existing research for facial expression recognition focused on valence-arousal estimation, facial action unit detection, and expression classification. We will introduce the latest related work of valence-arousal estimation study.

Many data are in laboratory settings. However, models that perform well on controlled conditions don't necessarily work well on uncontrolled ones. The ideal is still far from reality. Therefore, in-the-wild datasets come to exist. Kossaif et al. \cite{kossaifi2017afew} proposed a new dataset called AFEW-VA and found that geometric features performed well no matter what settings were. But it was unuseful for dynamic architecture since some of the clips in the dataset were too short to explore information between frame and frame. Barros et al. \cite{barros2018omg} proposed OMG dataset, collected from YouTube in real-world settings. The main keyword to select the videos was "monologue." Kollias et al. \cite{kollias2019deep} built a large-scale Aff-Wild dataset, collected from Youtube, and proposed deep convolutional and recurrent neural architecture, AﬀWildNet. A CNN extracted features while an RNN aimed to capture temporal information. Furthermore, their works not only got high performance on dimensional aspects but also for expression classification. 

Chang et al. \cite{chang2017fatauva} proposed an integrated deep learning framework that used the concept of applying the information of facial action unit detection to estimate valence-arousal intensity. They had shown that exploring the relationship between AUs and V-A was helpful for V-A research. Pan et al. \cite{pan2019deep} proposed a two-stream network to utilize effective facial features. The model contained CNN and LSTM. For temporal stream, the former extracted temporal features; the latter resolved the temporal relation between frames. For spatial stream similar to temporal one, the former extracted spatial features; the latter analyzed the spatial association between frames. Kim et al. \cite{kim2021contrastive} tackled regression problems with adversarial learning, which enabled the model to better understand complex emotion and achieved person-independent facial expression recognition. Also, they proposed a contrastive loss function and improved the performance effectively. This study proved the potential of adversarial learning instead of conventional methods on emotion recognition.

\begin{figure*}
  \centering
  \includegraphics[width=1\textwidth]{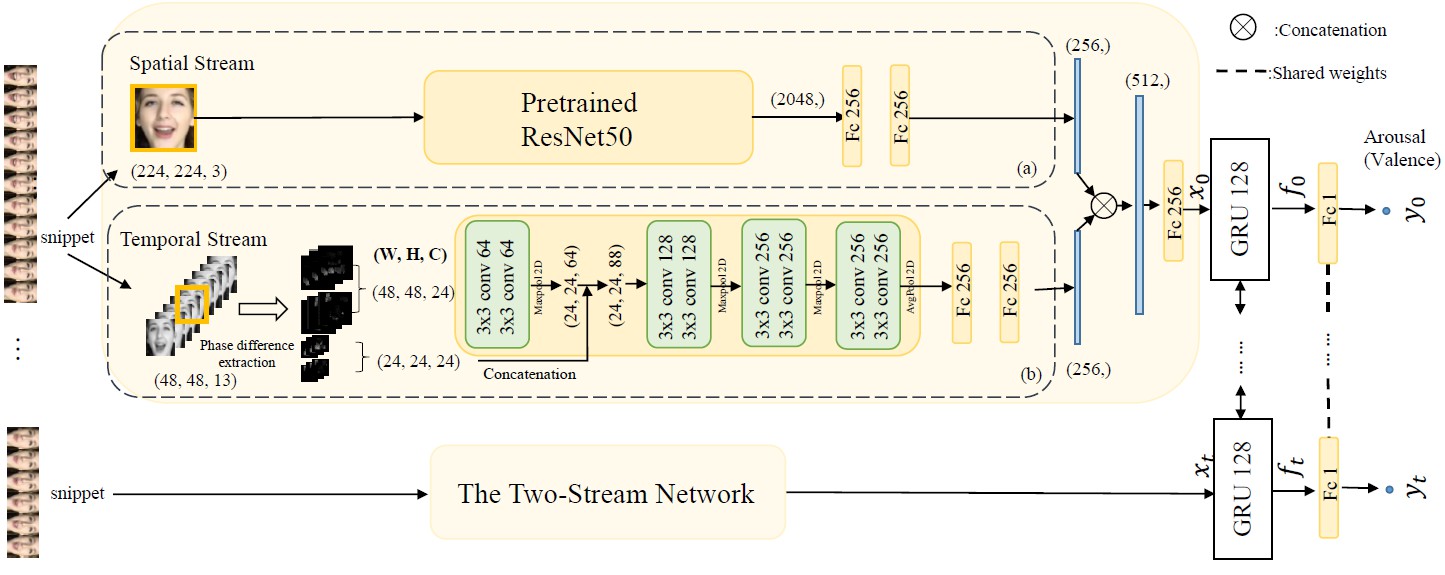}
  \caption{MIMAMO Net \cite{deng2020mimamo}}
\end{figure*}

\section{METHODOLOGY \& EXPERIMENTS}
\subsection{Dataset}
We only use the large-scale in-the-wild Aff-Wild2 dataset for our experiments since Deng et al. \cite{deng2020multitask} found that data balancing technique doesn't improve the performance of regression problem. This dataset contains 558 videos with frame-level annotations for valence-arousal estimation, facial action unit detection, and expression classification tasks. Our model focuses on estimating valence and arousal values, which take values in -1 to 1 and -5 represent no annotated values. In the VA set, there are 422 subjects with 1,932,935 images in the training and validation and 139 subjects with 714,986 images in the test. These cropped and aligned images were all provided by ABAW FG-2020 Competition organizers.


\subsection{Model}
In this research, we build our model based on MIMAMO Net \cite{deng2020mimamo}, which uses a two-stream network with the GRU model to integrating information about micro-motion and macro-motion. The inputs of the architecture separate into a spatial stream and temporal stream. It uses the pre-trained ResNet50 model to extract features and then fed pooling feature vector into fully connected layers to get the final feature vector in the spatial stream. While in the temporal stream, it utilizes time series of phase difference images to obtain the relationship between frames and then fed into the CNN network. The output of the two-stream network connects to the GRU model, which combines the information of the whole video to achieve frame-level predictions of valence and arousal values. 

To measure the agreement between the outputs of the model and the ground truth, it uses Concordance Correlation Coefficient (CCC) metrics as follow:
\[
CCC=\frac{2\rho\sigma_x\sigma_y}{\sigma^2_x+\sigma^2_y+(\mu_x-\mu_y)^2}
\]
where x and y are the predictions and annotations, $\mu_x$ and $\mu_y$ are the mean values, $\sigma^2_x$ and $\sigma^2_y$ are their variances, and $\rho$ is the correlation coefficient.

\subsection{Data Pre-processing}
We merge the training set and validation set and use the cross-validation method to acquire a more accurate estimate of model prediction performance. To apply the Aff-Wild2 dataset, we remove unannotated frames from the beginning and let the remaining ones match its annotation values. However, when testing, this approach may cause problems, such as missing data. We address those missing frames by using two different methods to deal with two situations. If disregarding frames are at the beginning of the video, we label -5 as the prediction. And if removed frames are not the case of above, we take the predicted value of the previous frame as its estimation.

\subsection{Results}
Our valence CCC and arousal CCC are 0.415 and 0.511 on the reselected validation set. We find that the performance on arousal is better than the performance on valence. Since arousal describes how active the person is, it should be more related to facial motion than valence. Therefore, it is reasonable that we get higher accuracy on arousal.

\section{Conclusion}
We have conducted valence-arousal estimation in Affwild2 dataset by using MIMAMO Net \cite{deng2020mimamo}. In the future, we plan to improve the MIMAMO Net to achieve better result.

\bibliographystyle{IEEEtran}
\bibliography{ref}

\begin{thebibliography}{10}
\providecommand{\url}[1]{#1}
\csname url@samestyle\endcsname
\providecommand{\newblock}{\relax}
\providecommand{\bibinfo}[2]{#2}
\providecommand{\BIBentrySTDinterwordspacing}{\spaceskip=0pt\relax}
\providecommand{\BIBentryALTinterwordstretchfactor}{4}
\providecommand{\BIBentryALTinterwordspacing}{\spaceskip=\fontdimen2\font plus
\BIBentryALTinterwordstretchfactor\fontdimen3\font minus
  \fontdimen4\font\relax}
\providecommand{\BIBforeignlanguage}[2]{{%
\expandafter\ifx\csname l@#1\endcsname\relax
\typeout{** WARNING: IEEEtran.bst: No hyphenation pattern has been}%
\typeout{** loaded for the language `#1'. Using the pattern for}%
\typeout{** the default language instead.}%
\else
\language=\csname l@#1\endcsname
\fi
#2}}
\providecommand{\BIBdecl}{\relax}
\BIBdecl

\bibitem{deng2020mimamo}
D.~Deng, Z.~Chen, Y.~Zhou, and B.~Shi, ``Mimamo net: Integrating micro-and
  macro-motion for video emotion recognition,'' in \emph{Proceedings of the
  AAAI Conference on Artificial Intelligence}, vol.~34, no.~03, 2020, pp.
  2621--2628.

\bibitem{kollias2020analysing}
D.~Kollias, A.~Schulc, E.~Hajiyev, and S.~Zafeiriou, ``Analysing affective
  behavior in the first abaw 2020 competition,'' in \emph{2020 15th IEEE
  International Conference on Automatic Face and Gesture Recognition (FG
  2020)(FG)}, pp. 794--800.

\bibitem{kollias2019expression}
D.~Kollias and S.~Zafeiriou, ``Expression, affect, action unit recognition:
  Aff-wild2, multi-task learning and arcface,'' \emph{arXiv preprint
  arXiv:1910.04855}, 2019.

\bibitem{kollias2019face}
D.~Kollias, V.~Sharmanska, and S.~Zafeiriou, ``Face behavior a la carte:
  Expressions, affect and action units in a single network,'' \emph{arXiv
  preprint arXiv:1910.11111}, 2019.

\bibitem{zafeiriou2017aff}
S.~Zafeiriou, D.~Kollias, M.~A. Nicolaou, A.~Papaioannou, G.~Zhao, and
  I.~Kotsia, ``Aff-wild: Valence and arousal ‘in-the-wild’challenge,'' in
  \emph{Computer Vision and Pattern Recognition Workshops (CVPRW), 2017 IEEE
  Conference on}.\hskip 1em plus 0.5em minus 0.4em\relax IEEE, 2017, pp.
  1980--1987.

\bibitem{kollias2017recognition}
D.~Kollias, M.~A. Nicolaou, I.~Kotsia, G.~Zhao, and S.~Zafeiriou, ``Recognition
  of affect in the wild using deep neural networks,'' in \emph{Computer Vision
  and Pattern Recognition Workshops (CVPRW), 2017 IEEE Conference on}.\hskip
  1em plus 0.5em minus 0.4em\relax IEEE, 2017, pp. 1972--1979.

\bibitem{barros2018omg}
P.~Barros, N.~Churamani, E.~Lakomkin, H.~Siqueira, A.~Sutherland, and
  S.~Wermter, ``The omg-emotion behavior dataset,'' in \emph{2018 International
  Joint Conference on Neural Networks (IJCNN)}.\hskip 1em plus 0.5em minus
  0.4em\relax IEEE, 2018, pp. 1--7.

\bibitem{kollias2019deep}
D.~Kollias, P.~Tzirakis, M.~A. Nicolaou, A.~Papaioannou, G.~Zhao, B.~Schuller,
  I.~Kotsia, and S.~Zafeiriou, ``Deep affect prediction in-the-wild: Aff-wild
  database and challenge, deep architectures, and beyond,'' \emph{International
  Journal of Computer Vision}, pp. 1--23, 2019.

\bibitem{kossaifi2017afew}
J.~Kossaifi, G.~Tzimiropoulos, S.~Todorovic, and M.~Pantic, ``Afew-va database
  for valence and arousal estimation in-the-wild,'' \emph{Image and Vision
  Computing}, vol.~65, pp. 23--36, 2017.

\bibitem{chang2017fatauva}
W.-Y. Chang, S.-H. Hsu, and J.-H. Chien, ``Fatauva-net: An integrated deep
  learning framework for facial attribute recognition, action unit detection,
  and valence-arousal estimation,'' in \emph{Proceedings of the IEEE conference
  on computer vision and pattern recognition workshops}, 2017, pp. 17--25.

\bibitem{pan2019deep}
X.~Pan, G.~Ying, G.~Chen, H.~Li, and W.~Li, ``A deep spatial and temporal
  aggregation framework for video-based facial expression recognition,''
  \emph{IEEE Access}, vol.~7, pp. 48\,807--48\,815, 2019.

\bibitem{kim2021contrastive}
D.~H. Kim and B.~C. Song, ``Contrastive adversarial learning for
  person-independent facial emotion recognition,'' 2021.

\bibitem{deng2020multitask}
D.~Deng, Z.~Chen, and B.~E. Shi, ``Multitask emotion recognition with
  incomplete labels,'' in \emph{2020 15th IEEE International Conference on
  Automatic Face and Gesture Recognition (FG 2020)(FG)}.\hskip 1em plus 0.5em
  minus 0.4em\relax IEEE Computer Society, 2020, pp. 828--835.

\end{thebibliography}

%








\end{document}